# Quantum-Inspired Edge Detection Algorithms Implementation using New Dynamic Visual Data Representation and Short-Length Convolution Computation

Artyom M. Grigoryan[1], Sos S. Agaian[2], Karen Panetta[3]

[1] Department of Electrical and Computer Engineering, The University of Texas at San Antonio, USA
[2] Computer Science Department, The College of Staten Island New York, USA
[3] Department of Electrical and Computer Engineering, Tufts University, Medford, MA 02155, USA

E-mail: amgrigoryan@utsa.edu

## Abstract

As the availability of imagery data continues to swell, so do the demands on transmission, storage and processing power. Processing requirements to handle this plethora of data is quickly outpacing the utility of conventional processing techniques. Transitioning to quantum processing and algorithms that offer promising efficiencies over conventional methods can address some of these issues. However, to make this transformation possible, fundamental issues of implementing real time Quantum algorithms must be overcome for crucial processes needed for intelligent analysis applications. For example, consider edge detection tasks which require time-consuming acquisition processes and are further hindered by the complexity of the devices used thus limiting feasibility for implementation in real-time applications. Convolution is another example of an operation that is essential for signal and image processing applications, where the mathematical operations consist of an intelligent mixture of multiplication and addition that require considerable computational resources. This paper studies a new paired transform-based quantum representation and computation of one-dimensional and 2-D signals convolutions and gradients. A new visual data representation is defined to simplify convolution calculations making it feasible to parallelize convolution and gradient operations for more efficient performance. The new data representation is demonstrated on multiple illustrative examples for quantum edge detection, gradients, and convolution. Furthermore, the efficiency of the proposed approach is shown on real-world images.

## 1. Introduction

Convolution is an essential operation in signal and image processing applications and is a fundamental tool in computer vision and intelligent systems. It is utilized predominantly in signal and image processing applications including edge detection [1]-[3], filtering [4,5], object/face recognition [7,8], convolution and neural networks (CNN) [9,12,33,35]. Convolution is the natural mathematical operation; a) performed by a linear and time-invariant system over its input signal and b) on two functions $f$ and $h$, producing a third function $f * h$ that is typically viewed as a modified version of one of the original functions, giving the area overlap between the two functions as a function of the amount that one of the original functions is translated, and c) that is the multiplication in the frequency domain. The math behind convolution is a clever mixture of multiplication and addition, which are the essential components of quantum amplitude arithmetic (QAA). The multiplication operation is straightforward because quantum computers use the architecture of the product of a unitary process. Conversely, the addition operation on amplitudes is not natural for quantum computers [10].

Quite a few quantum visual data (QVD) processing algorithms have been developed [36]. They utilize quantum-mechanics principles to overcome the limitations of conventional QVD processing procedures. It should be noted that the application, namely performing the transformation on qubits of many operations and algorithms that are used in a classic computer, is not a simple task. New algorithms must be described in terms of qubits, and therefore such algorithms are complex puzzles. Additionally, deep quantum learning remains challenging, due to barriers implementing nonlinearities with quantum unitarizes [31,32]. The well-known fast methods for computing the convolution are based on the discrete Fourier transform [14]-[17]. However, in quantum computation, the application of the quantum Fourier transform (QFT) [18]-[21] for calculating the convolution is a difficult task [29].

Considering the QFT and inverse QFT are: a) more efficient than their classical counterparts, and the FFT and inverse FFT are the cornerstones for the convolution b) can be used a quantum implementation of the filtering on a grayscale image, by using the QFT, two images and ideal filters, and the principle of the quantum oracle [22], c) can be done on a quantum state consisting of $N = 2^r$ complex values with complexity $O(\log 2N) = O(r^2)$, and d) can potentially be used for the convolution computation in parallel [29,38], we deduce that it is reasonable to construct quantum analogs of convolution algorithms that outperform their classical counterparts using QFT. Achieving these milestones in quantum convolution will open new pathways for many interesting applications including convolution in neural networks and image enhancement [11]-[13].

Since there is no physically realizable method to compute the normalized convolution or correlation of the coefficients of two quantum states [23], converting classical operations for quantum is a main challenge. However, our hypothesis is quantum convolution is solvable, and the schemes exist. In particular, the quantum convolution of a signal with a short-length impulse response of a system or filter should have a simple solution.

In this paper, we present an efficient implementation of one-dimensional (1-D) linear convolution in a quantum computer. The quantum representation of signals and images has many different forms [24]-[26], and the right choice of representation is the key in calculating the convolution. To demonstrate the usability of our method we use short 1-D convolutions and gradient operators with short-length masks of such as [1 -2 1], [1 2 2 1], [1 2 0 -2 -1], [1 2 -6 2 1], and [1 1 -4 1 1]. These operators are widely used in image processing, such as edge detection by different gradient operators [4]-[6]. In these examples, we show how to choose the quantum representation of the signal for computing the convolution; each case is unique and considered separately. A standard quantum representation is introduced for such convolutions and the quantum paired transform (QPT) [26,29] are used to calculate these convolutions together with gradients.

We also propose a method of quantum convolution which is described in detail on the example with 3 qubits; the general case can be considered similarly. Here, we note that the convolution of length $2^r$, or $r$-qubit convolution, can be sequentially separated by short convolutions [29]. Therefore, the presence of the schemes for short convolutions will make it possible to implement the calculation of convolution of the length $2^r$, $r \geq 3$. The case is considered, when the discrete Fourier transform of one of the signals is known and only the QFT of another signal is calculated. The frequency characteristics of many linear time invariant systems and filters are well known. Therefore, the considered method of convolution can be used for these systems in quantum computation. The Fourier transform method is very efficient when computing convolution in a classical computer; the convolution is reduced to multiplication. But it is this multiplication operation that is the most difficult step in quantum convolution using the Fourier transform. To overcome this obstacle, we suggest using an additional qubit and perform the corresponding permutation and prepare the quantum superposition of qubits for the inverse QFT. The example of the circuits for the low-pass and high-pass filters are given.

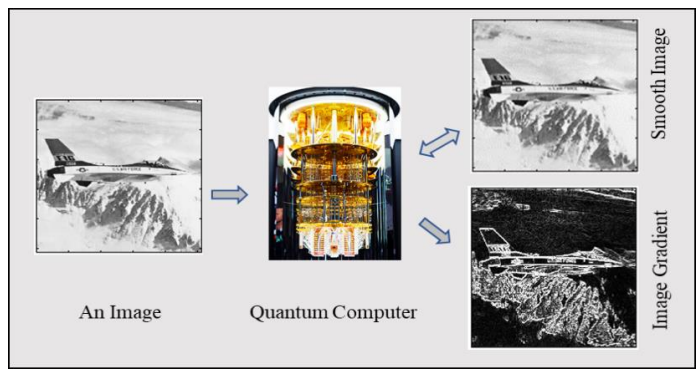

Fig. 1. Image gradient computation by a quantum computer.

The key contributions of this work are:
- A new paired transform-based quantum representation and computation of one-dimensional and 2-D signal convolutions and gradients.
- Simultaneous computation of a few convolutions and gradients (Fig. 1).
- Several illustrative 2-qubit and 3-qubits examples of quantum algorithms, including edge detection, gradients, and convolution algorithms.

The rest of the paper is organized in the following way. Section 2 describes the fundamental concepts of the qubit and quantum superposition of qubits in the standard computational base of states. Section 3 presents the method of 1-D convolution of the signal, which is written in detail for the short-length mask, by using the discrete paired transform. In Sections 4 and 5, a few examples of the convolution presenting the gradients, including the Sobel gradient operator, are described. The computer simulation of the measurements of results of the proposed method is illustrated on examples with images.

## 2. Basic Concepts of Qubits

In theory of quantum computation, the qubits are described by the superposition of states as elements of the vector subspace, which have the length equal one [10]. For instance, an individual qubit is described by a superposition

$$|\varphi\rangle = a_0|0\rangle + a_1|1\rangle = \begin{bmatrix} a_0 \\ a_1 \end{bmatrix},$$

where the coefficients $a$ and $b$ are amplitudes such that $|a_0|^2 + |a_1|^2 = 1$. Here, $|0\rangle$ and $|1\rangle$ denote the computational basis states $|0\rangle = [1,0]'$ and $|1\rangle = [0,1]'$. The qubit can be measured only once and the measurement results in only one state, $|0\rangle$ or $|1\rangle$ with probability $|a_0|^2$ or $|a_1|^2$, respectively. Two-qubit quantum superposition can be written as

$$|\varphi\rangle = a_0|00\rangle + a_1|01\rangle + a_2|10\rangle + a_3|11\rangle,$$

with the basis states $|00\rangle = [1,0,0,0]'$, $|01\rangle = [0,1,0,0]'$, $|10\rangle = [0,0,1,0]'$, and $|11\rangle = [0,0,0,1]'$. The amplitudes $a_k, k = 0:3$, define the probabilities $|a_k|^2$ of measurement the 2-qubit in one of the states $|k\rangle$. The operation $\otimes$ of the tensor product, or the Kronecker product, of vectors is used widely in quantum calculations. For instance, we can write

$$|01\rangle = |0\rangle \otimes |1\rangle = \begin{bmatrix} 1 \\ 0 \end{bmatrix} \otimes \begin{bmatrix} 0 \\ 1 \end{bmatrix}.$$

A signal $\{f_n; n = 0: (N-1)\}$ of length $N$, which can be considered a power of two, $N = 2^r$, can be map into the space of $r$-qubit superposition of states as

$$|\varphi(f)\rangle = \sum_{n=0}^{N-1} f_n |n\rangle,$$

where the basic states

$$|n\rangle = |n_{r-1} \ldots n_1 n_0\rangle \triangleq |n_{r-1}\rangle \otimes \ldots \otimes |n_1\rangle \otimes |n_0\rangle$$

are written by using the binary representation of $n$. The norm of the signal is considered to be 1. Otherwise, the above superposition must be normalized. Thus, the values of the signal are written into the amplitudes of the superposition and they define the probability of the measurement of the qubits. Many forms exist for quantum representation of 1-D and 2-D signals (for detail, see [24]- [26]).

For two signals $f$ and $h$ written in quantum algebra, the linear or aperiodic convolution is considered as the superposition

$$|\psi\rangle = \sum_{n=0}^{N-1} (f * h)_n |n\rangle.$$

Here, the problem is in calculating this superposition from the superpositions $|\varphi(f)\rangle$ and $|\varphi(h)\rangle$, by using only unitary transforms. As mentioned above, this task is considered unsolvable [23]. Also, during the calculation, the same values of the signal are used for the convolution at different points. This fact is not obstacle in traditional computation, when the desired values can be saved and used if needed. However, it is important to mention that no-cloning theorems prohibit the copying of qubits [30,31].

## 3. Method of 1-D Quantum Convolution

In this section, we present the method of calculation of convolutions in a quantum computer. The convolution is considered to be an operation over the amplitudes of a quantum superposition of the signal. Two steps are used in this method. First, a quantum representation of the convolution at each point is defined. Such a representation may be written in different ways and lead to different results in calculations. The distinguished property of the proposed method is in the fact that together with the given convolution, quantum computing allows for parallel computing other convolutions as well. Many of these additional convolutions or gradients also can be useful when processing signals. Therefore, signal and convolution quantum representations, and we are sure of this, need to be analyzed separately for each specific case. In the second step of the proposed method, the quantum paired transform is applied to parallel a few convolutions and gradients. The examples below describe the proposed method.

### 3.1. Convolution Quantum Representation

Let us consider a signal $\{f_n; n = 0: (N-1)\}$ of length $N = 2^r$, $r > 2$, and the following mask for the convolution: $M = [1\ \underline{2}\ 2\ 1]$. The underlined number shows the position of the center of the mask. The signal can be zero-padded to the length $N$. Such signal can be represented by $r$ qubits. We consider that the signal is periodic to simplify the calculation of normalizing coefficients in the quantum representation of signals. The convolution of the signal with this mask is calculated at each point $n$ by

$$y_n = (f * M)_n = f_{n-2} + 2f_{n-1} + 2f_n + f_{n+1}. \qquad (1)$$

These components of the convolution at point $n$ can be written into the vector $\mathbf{y}_n = (f_{n-2}, 2f_{n-1}, 2f_n, f_{n+1})'$. We do not yet know how the signal will be recorded into a real quantum computer. But we believe now that the preparation of such amplitudes of the 2-qubits will be possible with the help of a classical computer. Now, we consider the basis state $|n\rangle$ together with the following 2-qubit state superposition:

$$|y_n\rangle = \frac{1}{A}(f_{n-2}|00\rangle + 2f_{n-1}|01\rangle + 2f_n|10\rangle + f_{n+1}|11\rangle) \quad (2)$$

where the coefficient $A = \sqrt{f_{n-2}^2 + 4f_{n-1}^2 + 4f_n^2 + f_{n+1}^2}$. This two-qubit can be received from the two-qubit in the superposition of states

$$|q_n\rangle = \frac{1}{B}(f_{n-2}|00\rangle + f_{n-1}|01\rangle + f_n|10\rangle + f_{n+1}|11\rangle)$$

by using the transform with the diagonal matrix as shown

$$\begin{pmatrix} 1 & 0 & 0 & 0 \\ 0 & 2 & 0 & 0 \\ 0 & 0 & 2 & 0 \\ 0 & 0 & 0 & 1 \end{pmatrix} \begin{pmatrix} f_{n-2} \\ f_{n-1} \\ f_n \\ f_{n+1} \end{pmatrix} = y_n = \begin{pmatrix} f_{n-2} \\ 2f_{n-1} \\ 2f_n \\ f_{n+1} \end{pmatrix}.$$

The coefficient $B = B(n) = \sqrt{f_{n-2}^2 + f_{n-1}^2 + f_n^2 + f_{n+1}^2}$. The $(r+2)$-qubit state superposition of all convolution vectors is defined as

$$|\varphi\rangle = |\varphi(y)\rangle = \frac{1}{\sqrt{N}} \sum_{n=0}^{N-1} |n\rangle |y_n\rangle. \quad (3)$$

Here, $|n\rangle$ denotes the quantum computational basis states of $r$ qubits, and $|n\rangle|y_n\rangle = |n\rangle \otimes |y_n\rangle$. We call such a superposition of the convolution *a standard superposition*. This superposition requires two additional qubits. The superposition can also be considered as

$$|\psi\rangle = \frac{1}{C} \sum_{n=0}^{N-1} |n\rangle |(f_{n-2}|00\rangle + 2f_{n-1}|01\rangle + 2f_n|10\rangle + f_{n+1}|11\rangle), \quad (4)$$

where the coefficient $C = \sqrt{6(f_0^2 + f_1^2 + \cdots + f_{N-1}^2)}$. The difference between these two representations is in the measurement. When measuring the first $r$ qubits in the superposition in Eq. 3, we obtain two qubits in a superposition of states $|y_n\rangle$ with probability equal to $1/N$. The possibility of measuring the same 2 qubits in the superposition in Eq. 4 equals $1/C^2$. Note that $|y_n\rangle$ are the superpositions of two qubits at points $n$, and our goal is to calculate the vales $y_n$ of the convolution after processing $(r+2)$ qubits being in the superposition $|\varphi\rangle$ or $|\psi\rangle$.

Now, we process $|y_n\rangle$ in Eq. 2 by the 2-qubit paired transform. In matrix form, the 4-point discrete paired transform (DPT), $\chi_4'$, of a signal $x = \{x_0, x_1, x_2, x_3\}$ is calculated by [27]

$$\chi_4'[x] = \begin{bmatrix} 1 & 0 & -1 & 0 \\ 0 & 1 & 0 & -1 \\ 1 & -1 & 1 & -1 \\ 1 & 1 & 1 & 1 \end{bmatrix} \begin{bmatrix} x_0 \\ x_1 \\ x_2 \\ x_3 \end{bmatrix}. \quad (5)$$

The $4 \times 4$ matrix with determinant 8 in this equation can be considered as the orthogonal (unitary) matrix, after multiplying by the diagonal matrix as shown,

$$[\chi_4'] = \begin{bmatrix} \frac{1}{\sqrt{2}} & 0 & 0 & 0 \\ 0 & \frac{1}{\sqrt{2}} & 0 & 0 \\ 0 & 0 & \frac{1}{2} & 0 \\ 0 & 0 & 0 & \frac{1}{2} \end{bmatrix} \begin{bmatrix} 1 & 0 & -1 & 0 \\ 0 & 1 & 0 & -1 \\ 1 & -1 & 1 & -1 \\ 1 & 1 & 1 & 1 \end{bmatrix}$$

$$= \begin{bmatrix} \frac{1}{\sqrt{2}} & 0 & -\frac{1}{\sqrt{2}} & 0 \\ 0 & \frac{1}{\sqrt{2}} & 0 & -\frac{1}{\sqrt{2}} \\ \frac{1}{2} & -\frac{1}{2} & \frac{1}{2} & -\frac{1}{2} \\ \frac{1}{2} & \frac{1}{2} & \frac{1}{2} & \frac{1}{2} \end{bmatrix}. \quad (6)$$

The $2 \times 2$ matrix describes the operation of butterfly for the paired transform

$$A_2 = \frac{1}{\sqrt{2}} \begin{bmatrix} 1 & -1 \\ 1 & 1 \end{bmatrix} = \begin{bmatrix} 0 & 1 \\ 1 & 0 \end{bmatrix} \frac{1}{\sqrt{2}} \begin{bmatrix} 1 & 1 \\ 1 & -1 \end{bmatrix} = XH,$$

where $X$ is the NOT gate and $H$ is the 2×2 Hadamard matrix. The quantum circuit for the 2-qubit DPT is shown in Fig. 2 [29]. The input in this circuit is the 2-qubit state superposition

$$|\varphi_2\rangle = \frac{1}{A} \sum_{n=0}^{3} x_n |n\rangle, \quad A = \sqrt{x_0^2 + x_1^2 + x_3^2 + x_3^2},$$

where $|n\rangle$ denotes the basis state. The bullet on the line denotes the control qubit.

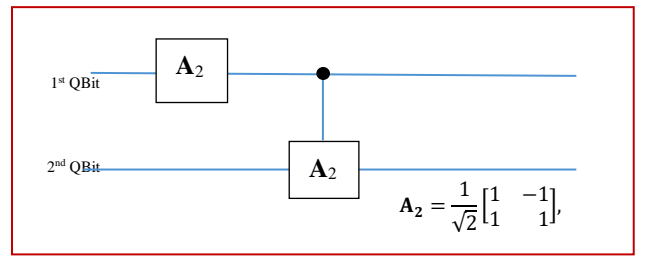

Fig. 2. The quantum circuit for the 2-qubit QPT.

For simplicity of calculations, we consider the 4-point DPT, as given in Eq. 5, i.e., which is defined by the matrix with the integer-valued coefficients. The paired transform of the vector $y_n$ at point $n$ is calculated by

$$\begin{bmatrix} 1 & 0 & -1 & 0 \\ 0 & 1 & 0 & -1 \\ 1 & -1 & 1 & -1 \\ 1 & 1 & 1 & 1 \end{bmatrix} \begin{bmatrix} f_{n-2} \\ 2f_{n-1} \\ 2f_n \\ f_{n+1} \end{bmatrix} = \begin{bmatrix} f_{n-2} - 2f_n \\ 2f_{n-1} - f_{n+1} \\ f_{n-2} - 2f_{n-1} + 2f_n - f_{n+1} \\ f_{n-2} + 2f_{n-1} + 2f_n + f_{n+1} \end{bmatrix}.$$

The result of the transform is the 2-qubit

$$|\chi'_4(y_n)\rangle = c_0|00\rangle + c_1|01\rangle + c_2|10\rangle + c_3|11\rangle \quad (7)$$

with the following amplitudes of the basic states: $c_0 = f_{n-2} - 2f_n$, $c_1 = 2f_{n-1} - f_{n+1}$, $c_2 = f_{n-2} - 2f_{n-1} + 2f_n - f_{n+1}$, and $c_3 = f_{n-2} + 2f_{n-1} + 2f_n + f_{n+1}$. These amplitudes should be normalized by the coefficient $\sqrt{c_0^2 + c_1^2 + c_2^2 + c_3^2}$.

One can see that (up to the constant 6) the amplitude $c_3$ is the value of the convolution, $y_n$, with the mask $[1\ 2\ 2\ 1]/6$,

$$c_3 = y_n = \frac{1}{6}[1\ 2\ 2\ 1]\begin{bmatrix}f_{n-2}\\f_{n-1}\\f_n\\f_{n+1}\end{bmatrix}. \quad (8)$$

The amplitude $c_2$ is the value of the convolution which is described by the mask $[-1\ 2\ -2\ 1]/3$ and represents the 4-level gradient operator,

$$c_2 = G_n(f) = \frac{1}{3}[1\ -2\ 2\ -1]\begin{bmatrix}f_{n-2}\\f_{n-1}\\f_n\\f_{n+1}\end{bmatrix}. \quad (9)$$

Figure 3 shows the circuit element for processing 2-qubit $|y_n\rangle$, by the 2-qubit paired transform. Thus, the paired transform allows for parallel computing two different convolutions, one of which is the gradient operation.

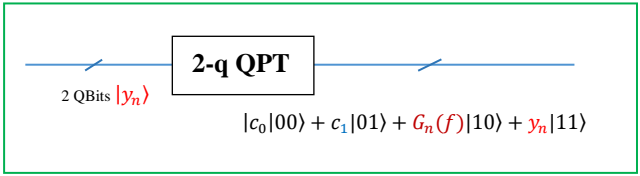

Fig. 3. The circuit element for processing the 2-qubit $|y_n\rangle$.

## 4. GRADIENT OPERATORS AND NUMERICAL SIMULATIONS

In this section, we describe a few experiments to show that the presented method works well. For example, we consider the quantum representation for the gradient which is described by the mask $M = [1\ -2\ 1]/2$. The gradient of the signal, or the convolution of the signal with this mask, is calculated at each point $n$ by

$$G_n(f) = (f * M)_n = [f_{n-1} - 2f_n + f_{n+1}]/2. \quad (10)$$

This operation can be written as

$$G_n(f) = [(f_{n-1} - f_n) + (f_{n+1} - f_n)]/2.$$

We define the following 2-qubit state superposition at point $n$:

$$|y_n\rangle = \frac{1}{A}(f_{n-1}|00\rangle - f_n|01\rangle + f_{n+1}|10\rangle - f_n|11\rangle) \quad (11)$$

and the vector $\mathbf{y}_n = (f_{n-2}, -f_n, f_{n+1}, f_n)'$. Here, the coefficient $A = A(n) = \sqrt{f_{n-1}^2 + 2f_n^2 + f_{n+1}^2}$. The $(r+2)$-qubit state superposition for the gradient of the signal is considered in standard form (3).

The following superposition also can be used:

$$|\psi\rangle = |\psi(y)\rangle = \frac{1}{C}\sum_{n=0}^{N-1}|n\rangle|y_n\rangle =$$

$$= \frac{1}{C}\sum_{n=0}^{N-1}|n\rangle|(f_{n-1}|00\rangle - f_n|01\rangle + f_{n+1}|10\rangle - f_n|11\rangle).$$

The coefficient $C = 2\sqrt{(f_0^2 + f_1^2 + \cdots + f_{N-1}^2)}$.

The 4-point discrete paired transform of the amplitudes of $|y_n\rangle$ is

$$\begin{bmatrix}1 & 0 & -1 & 0\\0 & 1 & 0 & -1\\1 & -1 & 1 & -1\\1 & 1 & 1 & 1\end{bmatrix}\begin{bmatrix}f_{n-1}\\-f_n\\f_{n+1}\\-f_n\end{bmatrix} = \begin{bmatrix}f_{n-1} - f_{n+1}\\0\\f_{n-1} + 2f_n + f_{n+1}\\f_{n-1} - 2f_n + f_{n+1}\end{bmatrix}. \quad (12)$$

Thus, the 2-qubit paired transform on the input $|y_n\rangle$ is the 2-qubit superposition

$$|\chi'_4(y_n)\rangle = c_0|00\rangle + c_1|01\rangle + c_2|10\rangle + c_3|11\rangle \quad (13)$$

with the following amplitudes of the states: $c_0 = f_{n-1} - f_{n+1}$, $c_1 = 0$, $c_2 = f_{n-1} + 2f_n + f_{n+1}$, and $c_3 = f_{n-1} - 2f_n + f_{n+1}$. These amplitudes of states should be normalized by the coefficient $A = \sqrt{c_0^2 + c_2^2 + c_3^2}$. The amplitude $c_3$ is the value of the gradient $G_n(f)$ at point $n$,

$$c_3 = c_3(n) = G_n(f) = \frac{1}{2}[1\ -2\ 1]\begin{bmatrix}f_{n-1}\\f_n\\f_{n+1}\end{bmatrix}. \quad (14)$$

Up to the factor 4, the amplitude $c_2$ equals the convolution of the signal at point $n$, when the mask is $[1\ 2\ 1]/4$,

$$c_2 = c_2(n) = \frac{1}{4}[1\ 2\ 1]\begin{bmatrix}f_{n-1}\\f_n\\f_{n+1}\end{bmatrix}. \quad (15)$$

Thus, in this example also the two-qubit paired transform allows for parallel calculation of the convolution and the gradient of the signal. It should be noted that, for parallel calculating these two convolutions with mask of length 3, four amplitudes of 2-qubits in (11) were determined in a certain way.

For illustration, we consider the above operations over each row of length 512 of the image 'jetplane.jpg' of size 512×512 pixels in Fig. 4(a). The images composed by $c_0(n)$, $c_2(n)$, and $c_3(n)$ coefficients are shown in parts (b), (c), and (d), respectively. The images in parts (b) and (d) are gradient images, and in part (c) the image is smoothed along the $X$ axis.

Fig. 4. (a) The original grayscale image, (b) the $c_0$-gradient image, (c) $c_2$-smooth image, and (d) $c_3$-gradient image.

We can model the process of measurements of all $(r+2)$ qubits $|\psi\rangle$ for this image, and consider the probability of measurement of the 2-qubit $|y_n\rangle$ in basis states $|00\rangle$, $|10\rangle$, and $|11\rangle$ according to the coefficients $|c_0|^2$, $|c_2|^2$, and $|c_3|^2$. The result of such a simulation on the classical computer is shown in Fig. 5 in part (b). For each row of the image, at each point $n \in \{1,2,\ldots,512\}$, the value of row-signal was taken randomly from the corresponding set of amplitudes $\{c_0(n), c_2(n), c_3(n)\}$. The edge points can be extracted by the threshold operation of the quantum bit sequence [35].

Fig. 5. (a) The 'jetplane' image and (b) the simulated measured image with $c_0$, $c_2$, and $c_3$-images.

## 5. NUMERICAL SIMULATIONS: SOBEL GRADIENT OPERATORS

This section focuses on the conventional image processing edge detection task, which is the perception of boundaries (intensity changes) between two neighboring regions. The brain uses this task: It has been shown that the brain processes visual information by responding to lines and edges with different neurons, which is an essential step in many pattern recognition tasks [34]. Edge detection methods use an image gradient by applying different types of filtering masks. We consider the quantum representation of the 5-level Sobel gradient operator with the mask $M = [\text{-}1\ \text{-}2\ \underline{0}\ 2\ 1]/3$. The gradient of the signal, or the convolution of the signal with this mask, is calculated at each point $n$ by

$$G_5(n) = (f * M)_n = \frac{[f_{n-2} + 2f_{n-1} - 2f_{n+1} - f_{n+2}]}{3}. \quad (16)$$

This operation can be written as the sum with 8 terms

$$G_5(n) = \frac{1}{3}[(f_{n-2} - f_n) + 2(f_{n-1} - f_n) + 2(f_n - f_{n+1}) + (f_n - f_{n+2})]$$

$$= \frac{1}{3}[(f_{n-2} - f_n) + 2(f_{n-1} - f_n) - 2(f_{n+1} - f_n) - (f_{n+2} - f_n)].$$

**A.** We consider the following quantum representation of the 3-qubit for the convolution at point $n$:

$$|y_n\rangle = \frac{1}{A}(f_{n-2}|0\rangle - f_n|1\rangle + 2f_{n-1}|2\rangle - 2f_n|3\rangle + 2f_n|4\rangle - 2f_{n+1}|5\rangle + f_n|6\rangle - f_{n+2}|7\rangle).$$

Here, the coefficient

$$A = \sqrt{f_{n-2}^2 + 4f_{n-1}^2 + 10f_n^2 + 4f_{n+1}^2 + f_{n+2}^2}.$$

The corresponding 8-dimensional vector is

$$\boldsymbol{y}_n = (f_{n-2}, -f_n, 2f_{n-1}, -2f_n, 2f_n, -2f_{n+1}, f_n, -f_{n+2})'.$$

The $(r+3)$-qubit state superposition for the gradient of the signal is considered in standard form (3). We can also use the following superposition:

$$|\psi\rangle = \frac{1}{C}\sum_{n=0}^{N-1}|n\rangle \, |(f_{n-2}|0\rangle - f_n|1\rangle + 2f_{n-1}|2\rangle - 2f_n|3\rangle + 2f_n|4\rangle - 2f_{n+1}|5\rangle + f_n|6\rangle - f_{n+2}|7\rangle)$$

with the coefficient $C = \sqrt{20(f_0^2 + f_1^2 + \cdots + f_{N-1}^2)}$.

The 8-point discrete paired transform is defined by the following orthogonal matrix [27,28]:

$$[\chi_8'] = \text{diag}\left\{\frac{1}{\sqrt{2}}, \frac{1}{\sqrt{2}}, \frac{1}{\sqrt{2}}, \frac{1}{\sqrt{2}}, \frac{1}{2}, \frac{1}{2}, \frac{1}{\sqrt{8}}, \frac{1}{\sqrt{8}}\right\} \times$$

$$\begin{bmatrix} 1 & 0 & 0 & 0 & -1 & 0 & 0 & 0 \\ 0 & 1 & 0 & 0 & 0 & -1 & 0 & 0 \\ 0 & 0 & 1 & 0 & 0 & 0 & -1 & 0 \\ 0 & 0 & 0 & 1 & 0 & 0 & 0 & -1 \\ 1 & 0 & -1 & 0 & 1 & 0 & -1 & 0 \\ 0 & 1 & 0 & -1 & 0 & 1 & 0 & -1 \\ 1 & -1 & 1 & -1 & 1 & -1 & 1 & -1 \\ 1 & 1 & 1 & 1 & 1 & 1 & 1 & 1 \end{bmatrix}.$$

Therefore, for simplicity of calculations, we consider the 8-point discrete paired transform, $\chi'_8$, which has the above matrix with the integer-valued coefficients 0 and $\pm 1$. The transform over amplitudes of the 3-qubit $|y_n\rangle$ equals

$$\chi'_8[y_n] = \begin{bmatrix} 1 & 0 & 0 & 0 & -1 & 0 & 0 & 0 \\ 0 & 1 & 0 & 0 & 0 & -1 & 0 & 0 \\ 0 & 0 & 1 & 0 & 0 & 0 & -1 & 0 \\ 0 & 0 & 0 & 1 & 0 & 0 & 0 & -1 \\ 1 & 0 & -1 & 0 & 1 & 0 & -1 & 0 \\ 0 & 1 & 0 & -1 & 0 & 1 & 0 & -1 \\ 1 & -1 & 1 & -1 & 1 & -1 & 1 & -1 \\ 1 & 1 & 1 & 1 & 1 & 1 & 1 & 1 \end{bmatrix} \begin{bmatrix} f_{n-2} \\ -f_n \\ 2f_{n-1} \\ -2f_n \\ 2f_n \\ -2f_{n+1} \\ f_n \\ -f_{n+2} \end{bmatrix}$$

$$= \begin{bmatrix} -2f_n + f_{n-2} \\ -f_n + 2f_{n+1} \\ -f_n + 2f_{n-1} \\ -2f_n + f_{n+2} \\ f_{n-2} - 2f_{n-1} + f_n \\ f_n - 2f_{n+1} + f_{n+2} \\ f_{n-2} + 2f_{n-1} + 6f_n + 2f_{n+1} + f_{n+2} \\ f_{n-2} + 2f_{n-1} - 2f_{n+1} - f_{n+2} \end{bmatrix}. \quad (17)$$

One can see that the last four components of the paired transform are the convolutions of the signal with the masks [1 -2 1], [1 -2 1], [1 2 6 2 1], and [-1 -2 0 2 1]. The first four outputs describe the signal plus gradients. Indeed,

$$c_0 = -2f_n + f_{n-2} = -[f_n + (f_n - f_{n-2})],$$
$$c_1 = -f_n + 2f_{n+1} = f_{n+1} + (f_{n+1} - f_n),$$
$$c_2 = -f_n + 2f_{n-1} = f_{n-1} + (f_{n-1} - f_n),$$
$$c_3 = -2f_n + f_{n+2} = -[f_n + (f_n - f_{n+2})].$$

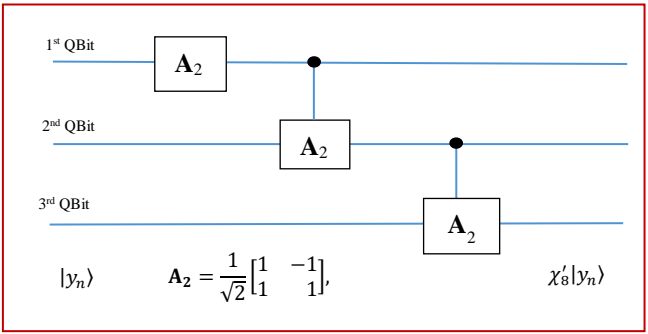

Fig. 6. The quantum circuit for the 3-qubit QPT.

The quantum circuit for computing the 3-qubit paired transform is shown in Fig. 6 [29]. The 3-qubit paired transform on the input $|y_n\rangle$ is the following 3-qubit state superposition:

$$|\chi'_8(y_n)\rangle = \frac{1}{C} \sum_{k=0}^{7} c_k |k\rangle.$$

Thus, when the first qubit is in state $|1\rangle$, the transform coefficients $c_4, c_5, c_6$, and $c_7$, or the amplitudes of the new superposition of states

$$|\chi'_8(y_n)\rangle_1 = \frac{1}{C_1}(c_4|00\rangle + c_5|01\rangle + c_6|10\rangle + c_7|11\rangle)$$

describe the different convolutions and gradients. Here, the normalizing coefficient $C_1 = \sqrt{c_4^2 + c_5^2 + c_6^2 + c_7^2}$.

Considering the 2-level Sobel gradient

$$G_2(n) = \frac{1}{2}[1 \ -2 \ 1] \begin{bmatrix} f_{n-1} \\ f_n \\ f_{n+1} \end{bmatrix},$$

we can write that the amplitudes $c_4$ and $c_5$ are the values of this gradient, which are calculated at two points, $(n-1)$ and $(n+1)$, i.e.,

$$c_4 = \frac{1}{2}[1 \ -2 \ 1] \begin{bmatrix} f_{n-2} \\ f_{n-1} \\ f_n \end{bmatrix} = G_2(n-1), \quad (18)$$

$$c_5 = \frac{1}{2}[1 \ -2 \ 1] \begin{bmatrix} f_n \\ f_{n+1} \\ f_{n+2} \end{bmatrix} = G_2(n+1).$$

The amplitude $c_6$ describes the convolution of the signal at point $n$, when the mask is [1 2 6 2 1]/12,

$$c_6 = C(n) = \frac{1}{12}[1 \ 2 \ 6 \ 2 \ 1] \begin{bmatrix} f_{n-2} \\ f_{n-1} \\ f_n \\ f_{n+1} \\ f_{n+2} \end{bmatrix}. \quad (19)$$

The last amplitude $c_7$ corresponds to the 5-level Sobel gradient of the signal at point $n$,

$$c_7 = G_5(n) = \frac{1}{3}[1 \ 2 \ 0 \ -2 \ -1] \begin{bmatrix} f_{n-2} \\ f_{n-1} \\ f_n \\ f_{n+1} \\ f_{n+2} \end{bmatrix}. \quad (20)$$

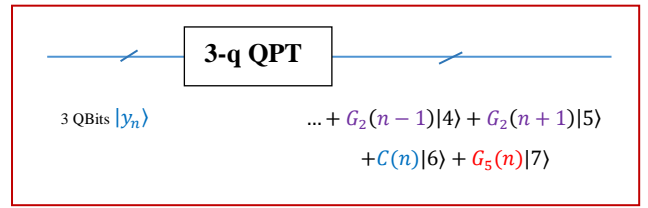

Fig. 7. The circuit element for processing the superposition $|y_n\rangle$.

Figure 7 shows the circuit element for processing the 2-qubit $|y_n\rangle$, by the 3-qubit paired transform over the input 3-qubit $|y_n\rangle$. This transform allows for parallel computing three different convolutions, one of which is the 5-level Sobel gradient transform. The 2-level gradient $G_2$ is calculated in two points, $(n-1)$ and $(n+1)$.

**B.** Now, we consider another quantum representation of the 3-qubit for the convolution at point $n$,

$$|y_n\rangle = \frac{1}{A}(f_{n-2}|0\rangle - f_n|1\rangle + 2f_{n-1}|2\rangle - 2f_n|3\rangle - 2f_{n+1}|4\rangle + 2f_n|5\rangle - f_{n+2}|6\rangle + f_n|7\rangle). \quad (21)$$

The corresponding 8-dimensional vector is

$$\boldsymbol{y}_n = (f_{n-2}, -f_n, 2f_{n-1}, -2f_n, -2f_{n+1}, 2f_n, -f_{n+2}, f_n)'.$$

The $(r+3)$-qubit state superposition for the gradient of the signal can be written as

$$|\varphi\rangle = |\varphi(y)\rangle = \frac{1}{\sqrt{N}}\sum_{n=0}^{N-1}|n\rangle|y_n\rangle.$$

The following superposition of states can also be considered:

$$|\psi\rangle = \frac{1}{C}\sum_{n=0}^{N-1}|n\rangle|(f_{n-2}|0\rangle - f_n|1\rangle + 2f_{n-1}|2\rangle - 2f_n|3\rangle - 2f_{n+1}|4\rangle + 2f_n|5\rangle - f_{n+2}|6\rangle + f_n|7\rangle).$$

Thus, if the first $r$ qubits are in the basis state $|n\rangle$, the next two qubits will be in the superposition $|y_n\rangle$. The 8-point discrete paired transform, $\chi'_8$, over the amplitudes of 2-qubit $|y_n\rangle$ equals

$$\chi'_8[\boldsymbol{y}_n] = \chi'_8 \begin{bmatrix} f_{n-2} \\ -f_n \\ 2f_{n-1} \\ -2f_n \\ -2f_{n+1} \\ 2f_n \\ -f_{n+2} \\ f_n \end{bmatrix} = \begin{bmatrix} f_{n-2} + 2f_{n+1} \\ -3f_n \\ 2f_{n-1} + f_{n+2} \\ -3f_n \\ f_{n-2} - 2f_{n-1} - 2f_{n+1} + f_{n+2} \\ 2f_n \\ f_{n-2} + 2f_{n-1} - 2f_{n+1} - f_{n+2} \\ f_{n-2} + 2f_{n-1} - 2f_{n+1} - f_{n+2} \end{bmatrix}.$$

Thus, the 3-qubit paired transform of the input qubit $|y_n\rangle$ is the following 3-qubit state superposition (up to a normalizing coefficient):

$$|\chi'_8(y_n)\rangle = \sum_{k=0}^{7} c_k|k\rangle. \quad (22)$$

Two amplitudes $c_6$ and $c_7$ describe the same convolution, namely, the 5-level Sobel gradient at point $n$,

$$c_6 = c_7 = G_5(n) = \frac{1}{3}[1 \quad 2 \quad 0 \quad -2 \quad -1]\begin{bmatrix} f_{n-2} \\ f_{n-1} \\ f_n \\ f_{n+1} \\ f_{n+2} \end{bmatrix}.$$

Therefore, if the first two qubits in the 3-qubit superposition $|\chi'_8(y_n)\rangle$ are in state $|1\rangle$ each, the measurement will result in the 5-level Sobel gradient at point $n$.

Up to the sign, the amplitude $c_4$ describes the convolution with the average and gradient operations,

$$c_4 = \frac{1}{2}[-1 \quad 2 \quad 0 \quad 2 \quad -1]\begin{bmatrix} f_{n-2} \\ f_{n-1} \\ f_n \\ f_{n+1} \\ f_{n+2} \end{bmatrix} =$$

$$= \frac{1}{2}(f_{n-1} + f_{n+1}) + \frac{1}{2}[-1 \quad 1 \quad 0 \quad 1 \quad -1]\begin{bmatrix} f_{n-2} \\ f_{n-1} \\ f_n \\ f_{n+1} \\ f_{n+2} \end{bmatrix}. \quad (23)$$

One can see that the representations of the convolution in parts **A** and **B** are different. Although the 5-level Sobel gradient is calculated in both representations, other convolutions and signal gradients are calculated.

**C. Other Gradient Operators**

The proposed method of quantum representation of short convolutions can also be used for other gradient operators [5,6]. For example, we consider the gradient operator with the mask $M = [1\ 1\ -4\ 1\ 1]/4$. The gradient of the signal $f_n$ at each point $n$ is calculated by

$$G_5(n) = (f * M)_n = [f_{n-2} + f_{n-1} - 4f_n + f_{n+1} + f_{n+2}]/4.$$

We consider the following quantum representation of the 3-qubit for the convolution at point $n$:

$$|y_n\rangle = \frac{1}{A}(f_{n-2}|0\rangle - f_n|1\rangle + f_{n-1}|2\rangle - f_n|3\rangle + f_{n+1}|4\rangle - f_n|5\rangle + f_{n+2}|6\rangle - f_n|7\rangle).$$

Here, the coefficient

$$A = A(n) = \sqrt{f_{n-2}^2 + f_{n-1}^2 + 4f_n^2 + f_{n+1}^2 + f_{n+2}^2}.$$

Therefore, the corresponding 8-dimensional vector at point $n$ is defined as

$$\boldsymbol{y}_n = (f_{n-2}, -f_n, f_{n-1}, -f_n, f_{n+1}, -f_n, f_{n+2}, -f_n)'.$$

The 8-point discrete paired transform of the convolution vector $\boldsymbol{y}_n$ equals

$$\chi'_8[\boldsymbol{y}_n] = \chi'_8 \begin{bmatrix} f_{n-2} \\ -f_n \\ f_{n-1} \\ -f_n \\ f_{n+1} \\ -f_n \\ f_{n+2} \\ -f_n \end{bmatrix} = \begin{bmatrix} f_{n-2} - f_{n+1} \\ 0 \\ f_{n-1} - f_{n+2} \\ 0 \\ f_{n-2} - f_{n-1} + f_{n+1} - f_{n+2} \\ 0 \\ f_{n-2} + f_{n-1} + 4f_n + f_{n+1} + f_{n+2} \\ f_{n-2} + f_{n-1} - 4f_n + f_{n+1} + f_{n+2} \end{bmatrix}.$$

Therefore, the 3-qubit paired transform on the input $|y_n\rangle$ is the 3-qubit state superposition

$$|\chi_8'(y_n)\rangle = \frac{1}{C} \sum_{k=0}^{7} c_k |k\rangle \qquad (24)$$

with the amplitudes $c_1 = c_3 = c_5 = 0$. Other five amplitudes represent a convolution and three gradient operations at point $n$. Indeed, $c_0 = c_0(n) = f_{n-2} - f_{n+1}$ $c_2 = c_2(n) = f_{n-1} - f_{n+2} = c_0(n+1)$, and (up to the factors 2, 8, and 2, respectively), and other amplitudes are

$$c_4 = c_4(n) = \frac{1}{2}[1 \quad -1 \quad 0 \quad 1 \quad -1]\begin{bmatrix}f_{n-2}\\f_{n-1}\\f_n\\f_{n+1}\\f_{n+2}\end{bmatrix},$$

$$c_6 = c_6(n) = \frac{1}{8}[1 \quad 1 \quad 4 \quad 1 \quad 1]\begin{bmatrix}f_{n-2}\\f_{n-1}\\f_n\\f_{n+1}\\f_{n+2}\end{bmatrix},$$

$$c_7 = c_7(n) = \frac{1}{2}[1 \quad 1 \quad -4 \quad 1 \quad 1]\begin{bmatrix}f_{n-2}\\f_{n-1}\\f_n\\f_{n+1}\\f_{n+2}\end{bmatrix}.$$

Thus, the 3-qubit paired transform allows for calculating three gradients and one convolution of the signal,

$$|\chi_8'(y_n)\rangle = \frac{1}{C}[\, c_0(n)|0\rangle + c_2(n)|2\rangle$$
$$+ c_4(n)|4\rangle + c_6(n)|6\rangle + c_7(n)|7\rangle\,]$$

where $C = \sqrt{c_0^2 + c_2^2 + c_4^2 + c_6^2 + c_7^2}$.

For the above 'jetplane' image processed by rows, Figure 8 shows the images composed by $c_0(n)$, $c_4(n)$, $c_6(n)$, and $c_7(n)$ coefficients in parts (a), (b), (c), and (d), respectively.

In this example, we also can model the process of measurements of all $(r+2)$ qubits $|\psi\rangle$ for the 'jetplane' image, and consider the probability of measurement of the 2-qubit $|y_n\rangle$ in five basis states $|000\rangle$, $|010\rangle$, $|100\rangle$, $|110\rangle$, and $|111\rangle$ according to the coefficients $|c_0|^2$, $|c_2|^2$, $|c_4|^2$, $|c_6|^2$, and $|c_7|^2$.

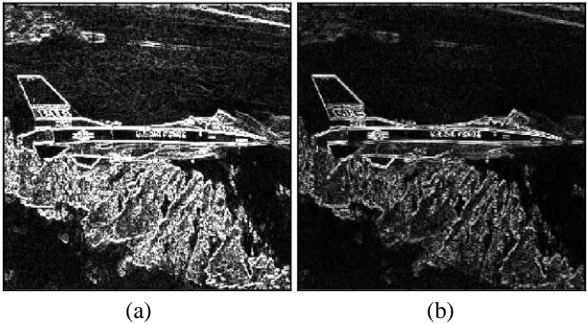

(a) (b)

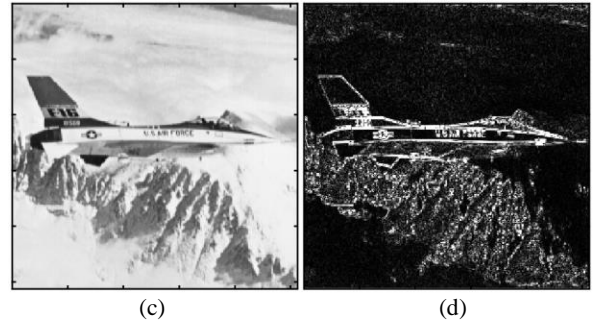

(c) (d)

Fig. 8. (a) The $c_0$-gradient image, (b) the $c_4$-gradient image, (c) $c_6$-smooth image, and (d) $c_7$-gradient image.

The result of such a simulation on the classical computer is shown in Fig. 9. For each row of the image, the values of row-signal were taken randomly from the corresponding set of amplitudes $\{c_0(n), c_2(n), c_4(n), c_6(n), c_7(n)\}$.

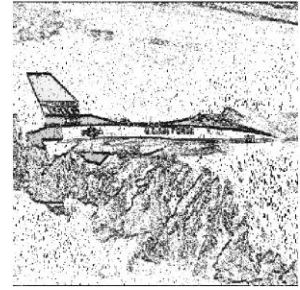

Fig. 9. The simulated measurement of the 'jetplane' image.

Figure 10 shows the grayscale image of Leonardo Da Vinci painting "Lady with Ermine" in part (a) and the result of computer simulation of measurements of qubits $|\psi\rangle$ along each row of this image in part (b).

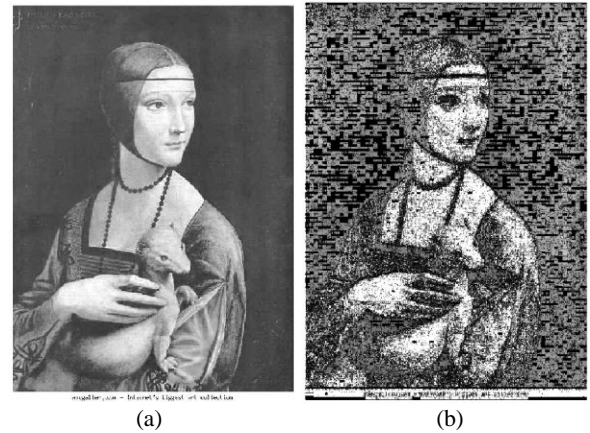

(a) (b)

Fig. 10. (a) The grayscale of [leonardo9.jpg] image of size 744×526 pixels (from http://www.abcgallery.com) and (b) the computer simulated measured image with $c_0, c_2, c_4, c_6$, and $c_7$-amplitudes.

The similar computer simulation of measurements for the grayscale image 'pepper' of size 512×512 pixels is shown in Fig. 11 in part (b).

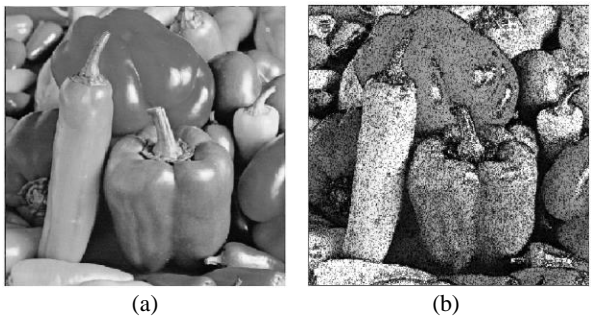

(a)                          (b)

Fig. 11. (a) The grayscale 'pepper' and (b) the simulated random image with $c_0, c_2, c_4, c_6$, and $c_7$-amplitudes

It should be noted that in comparison with the method of the convolution by the fast Fourier transform, which is used in traditional computations, the paired transform is much faster. It is the core of the discrete Fourier transform (DFT) [27,28]. The processing signals and images do not require the inverse transforms, as in the method of DFT. All images in the above figures are results of the direct paired transforms calculated along the rows of these 2-D signals. If we imagine that such a realization of the proposed convolution representation would be possible in a quantum computer, then a) the computation of convolution with gradients would be very efficient, b) quantum computers have the potential to resolve other challenges of computer vision and image processing applications, including multiscale analysis, machine learning, segmentation, pattern recognition, and coding.

## CONCLUSIONS

The quantum representation of convolutions is presented to calculate short-length convolutions and different gradients of the signal. For this, the quantum paired transform is used on amplitudes of the quantum states of the convolution at each point. It is essential to determine such a representation of the signal, which simplifies the procedure for calculating the convolution. Examples with convolutions and gradients with masks of length 3, 4, and 5 were described. These examples show that it is possible to build the circuits for the calculation of the quantum convolution. The presented method can be used for the calculation of other quantum short-length convolutions and gradients, as well. The only limitation of this method is that the impulse response of the filter or system is known. The paired transform is fast, binary, and is the kernel of the discrete Hadamard and Fourier transforms, which means that these two transforms can be decomposed by the sparse paired transforms [27,28]. Can these transforms be used instead of the discrete paired transform in the proposed method of convolution? We hypothesize the answer is "yes", but requires further research.

We strongly believe that the results presented in this article will stimulate further research in these fields. The proposed method of representation and computation can be generalized and used for other unitary transforms used in image and signal processing, including the Hadamard transforms [37].

## REFERENCES

[1] Robinson G.S.: "Edge detection by compass gradient masks," *Computer Graphics and Image Processing*, vol. 6, no. 5, pp. 492-501 (1977).

[2] Yuan, S., Venegas-Andraca, S.E., Wang, Y. et al., "Quantum image edge detection algorithm," Int J Theor Phys 58, 2823–2833 (2019).

[3] Fan, P., Zhou, RG., Hu, W. et al., "Quantum circuit realization of morphological gradient for quantum grayscale image," Int J Theor Phys 58, 415–435 (2019).

[4] Robinson G.S.: "Color edge detection," Proc. SPIE Symposium on Advances in Image Transmission Techniques, vol. 87, San Diego, CA (1976).

[5] Pratt W.K.: Digital Image Processing, 3rd Edition, John Wiley & Sons, Inc., (2001).

[6] Gonzalez R.C., Woods R.E.: Digital Image Processing, 2nd Edition, Prentice Hall, Upper Saddle River, New Jersey (2002).

[7] Ulyanov S., Petrov S.: "Quantum face recognition and quantum visual cryptography: Models and algorithms," *Electronic Journal, System Analysis in Science and Education*, no. 1, p. 17 (2012).

[8] M. Z. Khan, S. Harous, S. U. Hassan, M. U. Ghani Khan, R. Iqbal and S. Mumtaz, "Deep unified model for face recognition based on convolution neural network and edge computing," in *IEEE Access*, vol.7, pp. 72622-72633, 2019.

[9] Tan, RC., Liu, X., Tan, RG. *et al.*, "Cryptosystem for Grid Data Based on Quantum Convolutional Neural Networks and Quantum Chaotic Map," *Int J Theor Phys* 60, 1090–1102 (2021).

[10] Nielsen M., Chuang I.: Quantum Computation and Quantum Information, 2nd Ed., Cambridge UP (2001).

[11] V. Argyriou, T. Vlachos, R. Piroddi, "Gradient-adaptive normalized convolution," *IEEE Signal Processing Letters*, vol. 15, pp. 489-492 (2008).

[12] C. Cheng, K.K. Parhi, "Fast 2D convolution algorithms for convolutional neural networks," *IEEE Transactions on Circuits and Systems I: Regular Papers*, vol. 67, no. 5, pp. 1678-1691 (2020).

[13] Agaian S.S., Lentz K.P., Grigoryan A.M.: "Transform-based image enhancement algorithms," *IEEE Transaction on Image Processing,* vol. 10, no. 3, pp. 367-382 (2001).

[14] Cooley J.W., Tukey J.W.: "An algorithm the machine computation of complex Fourier series," *Math. Comput.,* vol. 9, no. 2, pp. 297–301 (1965).

[15] V.M. Amerbaev, R.A. Solovyev, A.L. Stempkovskiy, D.V. Telpukhov, "Efficient calculation of cyclic convolution by means of fast Fourier transform in a finite field," *Proceedings of IEEE East-West Design & Test Symposium (EWDTS 2014)*, Kiev, pp. 1-4 (2014).

[16] B. S. Paul S., A. X. Glittas, M. Sellathurai, G. Lakshminarayanan, "Reconfigurable 2, 3 and 5-point DFT processing element for SDF FFT architecture using fast cyclic convolution algorithm," *Electronics Letters*, vol. 56, no. 12, pp. 592-594 (2020).

[17] Blahut R.E.: Fast Algorithms for Digital Signal Processing. Addison-Wesley, Reading (1985).

[18] Cleve R., Watrous J.: "Fast parallel circuits for the quantum Fourier transform," in Proc. of IEEE 41st Annual Symposium on

Foundations of Computer Science, pp. 526-536, Redondo Beach, CA, USA (2000).

[19] Yoran N., Short A.: "Efficient classical simulation of the approximate quantum Fourier transform," *Phys. Rev. A* 76, 042321 (2007).

[20] Perez L.R., Garcia-Escartin J.C.: "Quantum arithmetic with the quantum Fourier transform," *Quantum Inf. Process*, vol. 16, p. 14 (2017).

[21] Grigoryan A.M., Agaian S.S.: "Paired quantum Fourier transform with $\log_2 N$ Hadamard gates," *Quantum Information Processing,* 18: 217, p. 26 (2019).

[22] Caraiman S., Manta V. I.: "Quantum image filtering in the frequency domain," *Advances in Electrical and Computer Engineering*, vol. 13, no. 3, pp. 77-84 (2013).

[23] Lomont C.: "Quantum convolution and quantum correlation are physically impossible," 2003, arXiv:quant-ph/:0309070. (Preprint)

[24] Yan F., Iliyasu A.M., Venegas-Andraca S.E.: "A survey of quantum image representations," *Quantum Inf. Process*. Vol. 15, no. 1, pp. 1–35 (2016).

[25] Yan F., Iliyasu A.M., Jiang Z.: "Quantum computation-based image representation, processing operations and their applications," *Entropy*, vol. 16, no. 10, pp. 5290-5338 (2014).

[26] Grigoryan A.M., Agaian S.S.: "New look on quantum representation of images: Fourier transform representation," *Quantum Information Processing,* 19:148, p. 26 (2020).

[27] Grigoryan A.M., Agaian S.S.: "Split manageable efficient algorithm for Fourier and Hadamard transforms," *IEEE Transaction on Signal Processing,* vol. 48, no. 1, pp. 172-183 (2000).

[28] Grigoryan A.M., Agaian S.S.: "Method of fast 1-D paired transforms for computing the 2-D discrete Hadamard transform," *IEEE Transactions on Circuits and Systems II*, vol. 48, no. 1 (2001).

[29] Grigoryan A.M.: "Resolution map in quantum computing: Signal representation by periodic pattens," *Quantum Information Processing*, 19:177, p. 21 (2020).

[30] W.K. Wootters, W.H. Zurek, A single quantum cannot be cloned, *Nature*, vol. 299, 802-803 (1982).

[31] D. Dieks, Communication by EPR Devices, Phys. Lett., vol. 92A, 271-272 (1982).

[32] M. Schuld, I. Sinayskiy, F. Petruccione, "The quest for a quantum neural network," *Quantum Information Processing*, vol. 13, no 11, pp. 2567–2586 (2014).

[33] M. Schuld, I. Sinayskiy, F. Petruccione, "An introduction to quantum machine learning," p. 19, arXiv:1408.7005 (2014).

[34] Xi-Wei Yao et al., "Quantum image processing and its application to edge detection: Theory and experiment," *Physical Review* X 7, 031041 (2017).

[35] I. Kerenidis, J. Landman, A, Prakash, "Quantum algorithms for deep convolutional neural network," p. 40 (2019), arXiv:1911.01117

[36] D. Emms et al., "Graph matching using the interference of discrete-time quantum walks," *Image and Vision Computing,* vol. 27, no. 7, pp. 934-949 (2009).

[37] Grigoryan A.M., Agaian S.S.: Multidimensional Discrete Unitary Transforms: Representation: Partitioning, and Algorithms, Marcel Dekker, Inc., 2003.